\documentclass[10pt,twocolumn,letterpaper]{article}

\usepackage{cvpr}
\usepackage{times}
\usepackage{epsfig}
\usepackage{graphicx}
\usepackage{amsmath}
\usepackage{amssymb}

\usepackage[pagebackref=true,breaklinks=true,letterpaper=true,colorlinks,bookmarks=false]{hyperref}

\usepackage{bbm}
\usepackage{bm}
\usepackage{amsmath}
\usepackage{amssymb}
\usepackage{dsfont}
\usepackage{graphicx} 
\usepackage{subfigure} 
\usepackage{algorithm}
\usepackage{algorithmic}
\usepackage{multirow}
\usepackage{enumitem}
\usepackage{array}
\newcolumntype{L}[1]{>{\raggedright\let\newline\\\arraybackslash\hspace{0pt}}m{#1}}
\newcolumntype{C}[1]{>{\centering\let\newline\\\arraybackslash\hspace{0pt}}m{#1}}
\newcolumntype{R}[1]{>{\raggedleft\let\newline\\\arraybackslash\hspace{0pt}}m{#1}}

\cvprfinalcopy 

\def\cvprPaperID{1480} 

\ifcvprfinal\fi
\begin{document}

\title{Visual Recognition by Counting Instances: A Multi-Instance Cardinality Potential Kernel}

\author{Hossein Hajimirsadeghi $\qquad$ Wang Yan $\qquad$ Arash Vahdat $\qquad$ Greg Mori\\
School of Computing Science, Simon Fraser University, Canada\\
{\tt\small hosseinh@sfu.ca, wyan@sfu.ca, avahdat@sfu.ca, mori@cs.sfu.ca}
}

\maketitle

\begin{abstract}
  Many visual recognition problems can be approached by counting instances. To determine whether an event is present in a long internet video, one could count how many frames seem to contain the activity.  Classifying the activity of a group of people can be done by counting the actions of individual people.  Encoding these cardinality relationships can reduce sensitivity to clutter, in the form of irrelevant frames or individuals not involved in a group activity. Learned parameters can encode how many instances tend to occur in a class of interest.  To this end, this paper develops a powerful and flexible framework to infer any cardinality relation between latent labels in a multi-instance model. Hard or soft cardinality relations can be encoded to tackle diverse levels of ambiguity. Experiments on tasks such as human activity recognition, video event detection, and video summarization demonstrate the effectiveness of using cardinality relations for improving recognition results.
\end{abstract}

\section{Introduction}
A number of visual recognition problems involve examining a set of instances, such as the people in an image or frames in a video.  For example, in group activity recognition (e.g.~\cite{choi09}) the prominent approach to analyzing the activity of a group of people is to look at the actions of individuals in a scene.  A number of impressive methods have been developed for modeling the {\em structure} of a group activity~\cite{lan12,choi12,amer12}, capturing spatio-temporal relations between people in a scene.  However, these methods do not directly consider cardinality relations about the {\em number} of people that should be involved in an activity.  These cardinality relations vary per activity.  An activity such as a fall in a nursing home~\cite{lan12} is different in composition from an activity such as queuing~\cite{choi12}, involving different numbers of people (one person falls, many people queue).  Further, clutter, in the form of people in a scene performing unrelated actions, confounds recognition algorithms.  In this paper we present a framework built on a latent structured model to encode these cardinality relations and deal with the ambiguity or clutter in the data.

Another example is unconstrained internet video analysis. Detecting events in internet videos~\cite{over11} or determining whether part of of a video is {\em interesting}~\cite{gygli14} are challenging for many reasons, including temporal clutter -- videos often contain frames unrelated to the event of interest or that are difficult to classify.  Two broad approaches exist for video analysis, either relying on holistic bag-of-words models or building temporal models of events.  Again, successful methods for modeling temporal structure exist (e.g.~\cite{gupta09_cvpr,TianSS13,tang12,vahdat13}).  Our method builds on these successes, but directly considers cardinality relations, counting how many frames of a video appear to contain a class of interest, and using soft and intuitive constraints such as ``the more, the better" to enhance recognition.

\begin{figure*}[tbh]
\centering
\subfigure[What is the collective activity?]{
\includegraphics[width=0.25\linewidth]{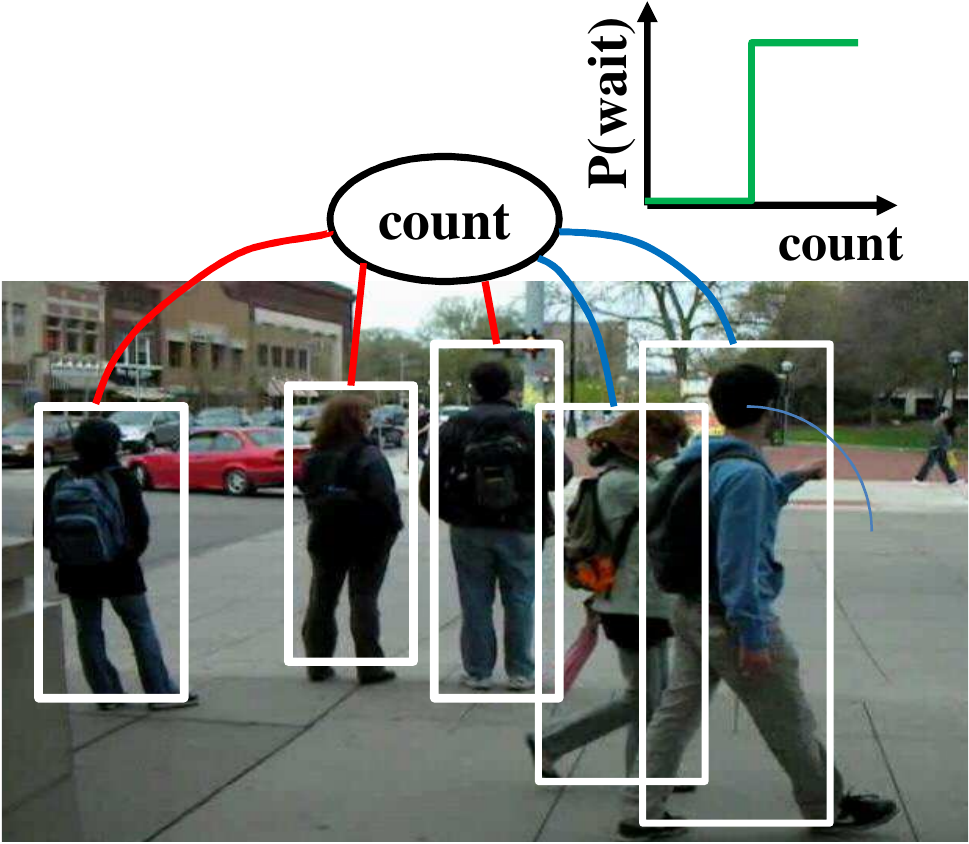}
\label{subfig:actvity}
}
\subfigure[What is this video about?]{
\includegraphics[width=0.35\linewidth]{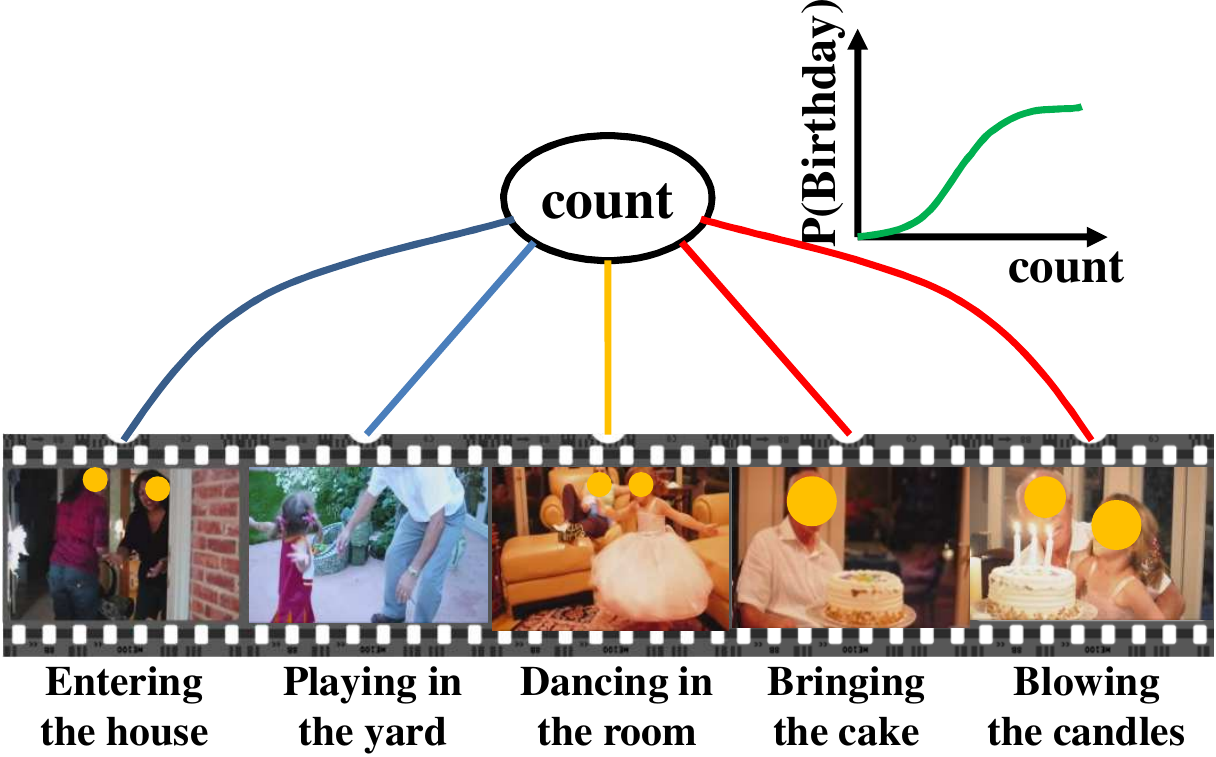}
\label{subfig:event-detection}
}
\subfigure[Is this video segment interesting?]{
\includegraphics[width=0.20\linewidth, angle=90]{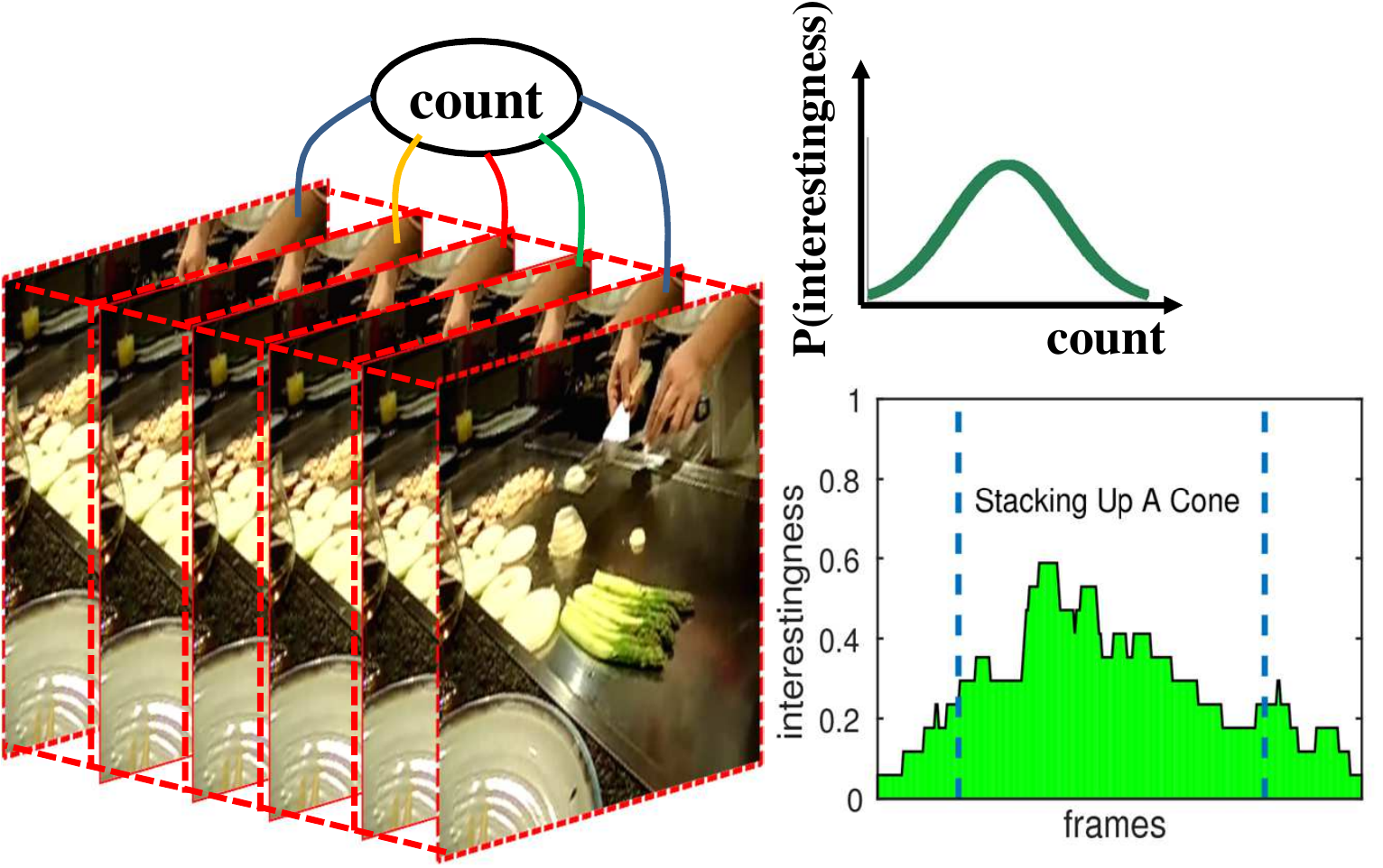}
\label{subfig:summarization}
}
\caption{Encoding cardinality relations can improve visual recognition. \subref{subfig:actvity} An example of collective activity recognition. Three people are waiting, and two people are walking (passing by in the street). Using only spatial relations, it is hard to infer what the dominant activity is, but encoding the cardinality constraint that the collective activity tends to be the majority action helps to break the tie and favor ``waiting" over ``walking". \subref{subfig:event-detection} A ``birthday party" video from the TRECVID MED11 dataset~\cite{over11}. Some parts of the video are irrelevant to birthdays and some parts share similarity with other events such as ``wedding". However, encoding soft cardinality constraints such as ``the more relevant parts, the more confident decision", can enhance event detection. \subref{subfig:summarization} A video from the SumMe summarization dataset~\cite{gygli14}. The left image shows an important segment, where the chef is stacking up a cone. The right image shows the human-judged interesting-ness score of each frame. Even based on human judgment, not all parts of an important segment are equally interesting. Due to uncertainty in labeling the start and end of a segment, the cardinality potential might be non-monotonic.}
\label{fig:intro}
\end{figure*}

Fig.~\ref{fig:intro} shows an overview of our method.  We encode our intuition about these counting relations in a multiple instance learning framework. In multiple instance learning, the input to the algorithm is a set of labeled {\em bags} containing {\em instances}, where the instance labels are not given. We approach this problem by modeling the bag with a probabilistic latent structured model. Here, we highlight the major contributions of this paper.
\begin{itemize}[leftmargin=*]
\item \textbf{Showing the importance of cardinality relations for visual recognition.} We show in different applications that encoding cardinality relations, either hard (e.g. \emph{majority}) or soft (e.g. \emph{the more, the better}), can help to enhance recognition performance and increase robustness against labeling ambiguity.
\vspace{-0.05in}
\item \textbf{A kernelized framework for classification with cardinality relations.} We use a latent structured model, which can easily encode any type of cardinality constraint on instance labels. A novel kernel is defined on these probabilistic models. We show that our proposed kernel method is effective, principled, and has efficient and exact inference and learning methods.
\end{itemize}

\section{Related Work}
\label{sec:rw}
This paper presents a novel model for cardinality relations in visual recognition, in particular for the analysis of video sequences.  Existing video analysis methods generally focus on structured spatio-temporal models, complementary to our proposed approach.  For instance, pioneering work was done by Gupta et al.~\cite{gupta09_cvpr} in analyzing structured videos by creating ``storyline'' models populated from AND-OR graph representations.  Related models have proven effective at analyzing scenes of human activity more broadly in work by Amer et al.~\cite{amer12}.  A series of recent papers has focused on the problem of group activity recognition, inferring an activity that is performed by a set of people in a scene.  Choi et al.~\cite{choi09,choi12}, Lan et al.~\cite{lan12}, and Khamis et al.~\cite{khamis-eccv2012} devised models for spatial and temporal relations between the individuals involved in a putative interaction.  Zhu et al.~\cite{ZhuNR13} consider contextual relations between humans and objects in a scene to detect interactions of interest.  The structural relations exploited by these methods are a key component of activity understanding, but present different information from the cardinality relations we study.

Analogous approaches have been studied for ``unconstrained'' internet video analysis.  Methods to capture the temporal structure of high-level events need to be robust to the presence of irrelevant frames.  Successful models include Tian et al.~\cite{TianSS13} and Niebles et al.~\cite{niebles10_eccv}, who extend latent variable models in the temporal domain.  Tang et al.~\cite{tang12} develop hidden Markov models with variable duration states to account for the temporal length of action segments.  Vahdat et al.~\cite{vahdat13} compose a test video with a set of kernel matches to training videos.
Tang et al.~\cite{tang13combining} effectively combine informative subsets of features extracted from videos to improve event detection. Bojanowski et al.~\cite{BojanowskiLBLPSS14} label videos with sequences of low-level actions. Pirsiavash and Ramanan~\cite{PirsiavashR14} develop stochastic grammars for understanding structured events. Xu et al.~\cite{xu14event} propose a feature fusion method based on utilizing related exemplars for event detection. Lai et al.~\cite{lai14} apply multiple instance learning to video event detection by representing a video as multi-granular temporal video segments.  Our work is similar in spirit, but contributes richer cardinality relations and more powerful kernel representations; empirically we show these can deliver superior performance.

The continued increase in the amount of video content available has rendered the summarization of unconstrained internet videos an important task. Kim et al.~\cite{KimSX14} build structured storyline-type representations for the events in a day.  Khosla et al.~\cite{KhoslaHLS13} use web images as a prior for selecting good summaries of internet videos. Popatov et al.~\cite{PotapovDHS14} learn the important components of videos of high-level events.  Gygli et al.~\cite{gygli14} propose a benchmark dataset for measuring interesting-ness of video clips and explore a set of high-level semantic features along with superframe segmentation for detecting interesting video clips.  We demonstrate that our cardinality-based methods can be effective for this task as well, scoring a clip by the number of interesting frames it contains.

\subsection{Multi-Instance Learning}
We develop an algorithm based on multiple instance learning, where an input example consists of a bag of instances, such as a video represented as a bag of frames.
The traditional assumption is that a bag is positive if it contains at least one positive instance, while in a negative bag all the instances are negative.  However, this is a very weak assumption, and recent work has developed advanced algorithms with different assumptions~\cite{li11text, hajimirsadeghi12, hajimirsadeghi13, lai14}.

For example, Li et al.~\cite{li11text} formulated a prior on the number of positive instances in a bag, and used an iterative cutting plane algorithm with heuristics to approximate the resultant learning problem.  Yu et al.~\cite{yu13} proposed $\propto$SVM for learning from instance proportions, and showed promising results on video event recognition~\cite{lai14}.
Our work improves on this approach by permitting more general cardinality relations with an efficient and exact training scheme.

Our approach models a bag of instances with a probabilistic model with a cardinality-based clique potential between the instance labels. This cardinality potential facilitates defining any cardinality relations between the instance labels and efficient and exact solutions for both maximum a posteriori (MAP) and sum-product inference~\cite{gupta07, tarlow12}. For example, Hajimirsadeghi et al.~\cite{hajimirsadeghi13} used cardinality-based models to embed different ratio-based multiple instance assumptions.  Here we extend these lines of work by developing a novel kernel-based learning algorithm that enhances classification performance.

Kernel methods for multiple instance learning include G{\"a}rtner et al.'s~\cite{gartner02} MI-Kernel, which is obtained by summing up the instance kernels between all instance pairs of two bags. Hence, all instances of a bag contribute to bag classification equally, although they are not equally important in practice. To alleviate this problem, Kwok and Cheung~\cite{kwok07} proposed marginalized MI-Kernel. This kernel specifies the importance of an instance pair of two bags according to the consistency of their probabilistic instance labels. In our work, we also use the idea of marginalizing joint kernels, but we propose a unified framework to combine instance label inference and bag classification within a probabilistic graph-structured kernel.

\section{Proposed Method: Cardinality Kernel}
\label{sec:method}

\begin{figure*}[tbh]
\centering
\subfigure[Preparing Training Data as Positive and Negative Bags of Instances]{
\includegraphics[width=0.25\linewidth]{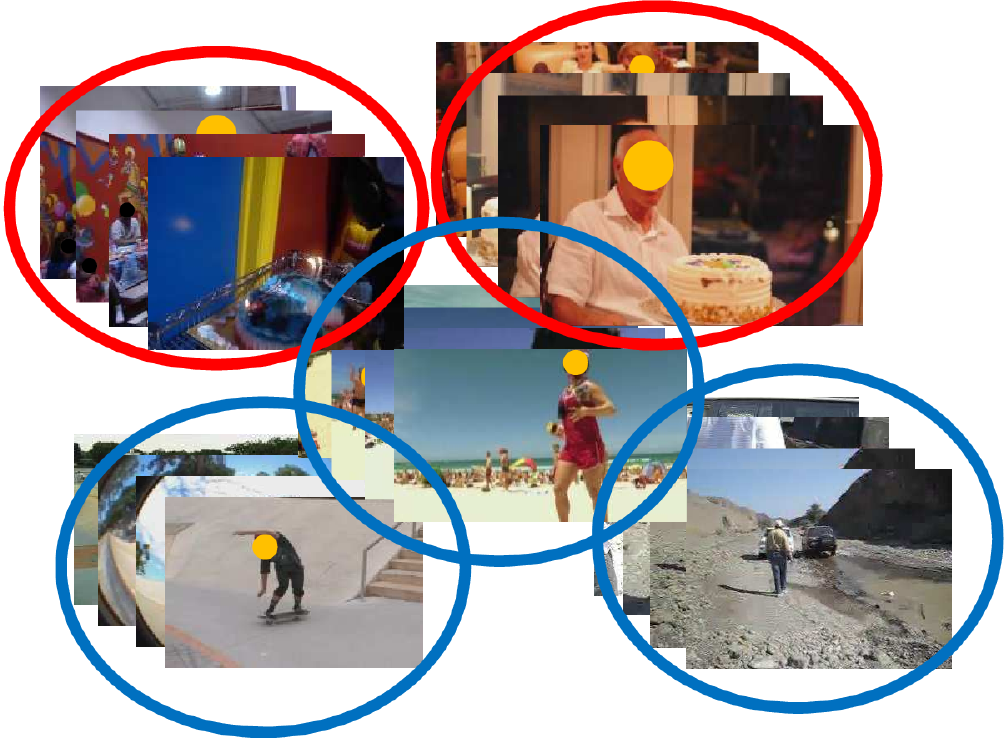}
\label{subfig:bags}
}
\quad
\subfigure[Learning the Cardinality Model]{
\includegraphics[width=0.18\linewidth]{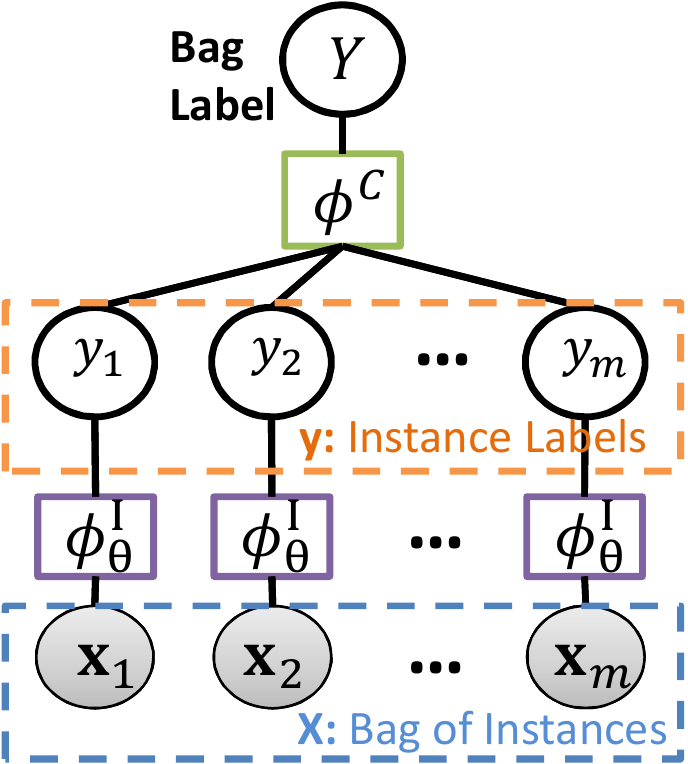}
\label{subfig:learning}
}
\quad
\subfigure[Computing Cardinality Kernels for all Bag Pairs]{
\includegraphics[width=0.23\linewidth]{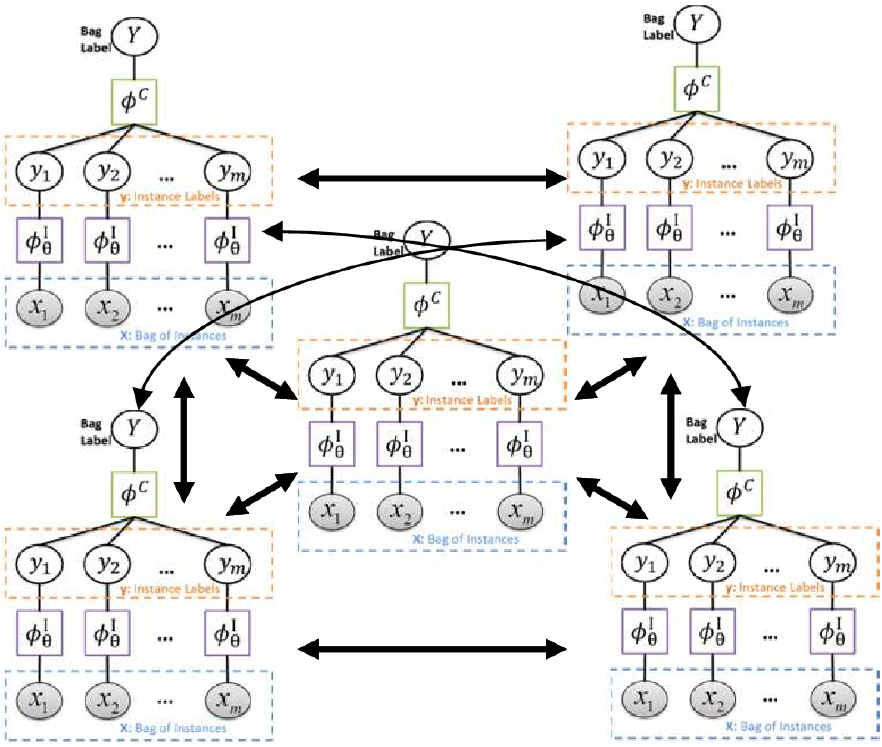}
\label{subfig:kernel}
}
\quad
\subfigure[Using a Kernel Method to Train a Classifier]{
\centering
\includegraphics[width=0.18\linewidth]{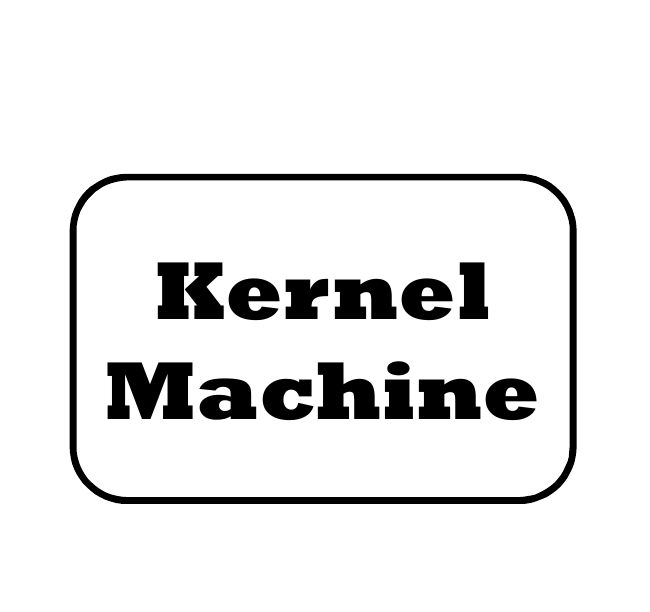}
\label{subfig:classifier}
}
\caption{The high-level scheme of the proposed kernel method for bag classification.}
\label{fig:method}
\end{figure*}

We propose a novel kernel for modeling cardinality relations, counting instance labels in a bag -- for example the number of people in a scene who are performing an action.  We start with a high-level overview of the method, following the depiction in Fig.~\ref{fig:method}.

The method operates in a multiple instance setting, where the input is bags of instances, and the task is to label each bag.  For concreteness, Fig.~\ref{fig:method}(a) shows video event detection.  Each video is a bag comprised of individual frames.  The goal is to label a video according to whether a high-level event of interest is occurring in the video or not.  Temporal clutter, in the form of irrelevant frames, is a challenge.  Some frames may be directly related to the event of interest, while others are not.  

Fig.~\ref{fig:method}(b) shows a probabilistic model defined over each video.  Each frame of a video can be labeled as containing the event of interest, or not.  Ambiguity in this labeling is pervasive, since the low-level features defined on a frame are generally insufficient to make a clear decision about a high-level event label.  The probabilistic model handles this ambiguity and a counting of frames -- parameters encode the appearance of low-level features and the intuition that more frames relevant to the event of interest makes it more likely that the video as a whole should be given the event label.  

A kernel is defined over these bags, shown in Fig.~\ref{fig:method}(c).  Kernels compute a similarity between any two videos.  In our case, this similarity is based on having similar cardinality relations, such as two videos having similar counts of frames containing an event of interest.  Finally, this kernel can be used in any kernel method, such as an SVM for classification, Fig.~\ref{fig:method}(d).

\subsection{Cardinality Model}
\label{subsec:cpm}
A cardinality potential is defined in terms of counts of variables which take some particular values. For example, with binary variables, it is defined in terms of the number of positively and negatively labeled variables.
Given a set of binary random variables $\mathbf{y} = \{y_1, y_2, \cdots, y_m\}$ ($y_i \in \{0, 1 \}$), the cardinality potential model is described by the joint probability
\begin{equation}\label{eq:scp}
 P(\mathbf{y}) = \frac{ C( \sum_i{y_i} ) \prod_i{\exp{(\varphi_i y_i)}} }{ \sum_{\mathbf{y}}{C( \sum_i{y_i} ) \prod_i{\exp{(\varphi_i y_i)}}} } ,
\end{equation}
which consists of one cardinality potential $C(\cdot)$ over all the variables and unary potentials $\exp{(\varphi_i y_i)}$ on features $\varphi_i$ on each single variable. Maximum a posteriori (MAP) inference of this model is straight-forward and takes $O\left(m \log{m} \right)$ time \cite{gupta07}. Sum-product inference is more involved, but efficient algorithms exist~\cite{tarlow12}, computing all marginal probabilities of this model in $O\left(m \log^2m \right)$ time.

In problems with multiple instances, there are assumptions or constraints which are defined on the counts of instance labels. For example, the standard multi-instance assumption states that at least one instance in a positive bag is positive. So, it is intuitive that these constraints can be modeled by a cardinality potential over the instance labels. This modeling helps to have exact and efficient solutions for MIL problems, using existing state-of-the-art inference and learning algorithms.

Using this cardinality potential model as the core, a probabilistic model of the likelihood of a bag of instances $\mathbf{X} = \lbrace \mathbf{x}_{1}, \mathbf{x}_{2}, \cdots, \mathbf{x}_{m} \rbrace$ with the bag label $Y \in \lbrace -1, +1 \rbrace$ and the instance labels $\mathbf{y}$ with model parameters $\bm{\theta}$, is built (c.f.~\cite{tarlow12}):
\begin{equation}\label{eq:bag-l}
 P(Y, \mathbf{y} | \mathbf{X}; \bm{\theta}) \propto \phi^C(Y, \mathbf{y}) \prod_i{\phi^I_{\bm{\theta}}(\mathbf{x}_i, y_i)}.
\end{equation}


\noindent A graphical representation of the model is shown in Fig.~\ref{fig:method}(b). In our framework, we call this the \emph{``Cardinality Model"}, and the details of its components are described as follows:

\noindent \textbf{Cardinality clique potential $\phi^C(Y, \mathbf{y})$:} a clique potential over all the instance labels and the bag label. This is used to model multi-instance or label proportion assumptions and is formulated as $ \phi^C(Y, \mathbf{y}) = C^{(Y)}(\sum_i{y_i})$. $C^{(+1)}$ and $C^{(-1)}$ are cardinality potentials for positive and negative bag labels, and in general could be expressed by any cardinality function.
In this paper we work with the ``normal" model in \eqref{eq:normalPot} and the ``ratio-constrained" model in \eqref{eq:majorityPot}.
\begin{equation}\label{eq:normalPot}
\begin{split}
 & C^{(+1)}(c) = \exp{\left( -(\frac{c}{m} - \mu)^2 / 2\sigma^2 \right)} \\
 & C^{(-1)}(c) = \exp{\left( -(\frac{c}{m})^2 / 2\sigma^2 \right)}.
\end{split}
\end{equation}
\begin{equation}\label{eq:majorityPot}
\begin{split}
 & C^{(+1)}(c) = \mathbbm{1}(\frac{c}{m} >= \rho)\\
 & C^{(-1)}(c) = \mathbbm{1}(\frac{c}{m} < \rho).
\end{split}
\end{equation}
The parameter $\mu$ in the normal model or $\rho$ in the ratio-constrained model controls the proportion of positive labeled instances in a bag. The Normal model does not impose hard constraints on the number of positive instances, and consequently a positive bag can have any proportion of positive instances but it is more likely to be around $\mu$. On the other hand, the ratio-constrained model makes a hard constraint, assuming a bag must have at least a certain ratio ($\rho$) of positive instances.

\noindent \textbf{Instance-label potential $\phi^I_{\bm{\theta}}(\mathbf{x}_i, y_i)$:} represents the potential between each instance and its label.  Essentially, this potential describes how likely it is for an instance (e.g.\ video frame) to receive a certain label (e.g.\ relevant or not to an event).  It is parameterized as:
\begin{equation}\label{eq:il-Pot}
\phi^I_{\bm{\theta}}(\mathbf{x}_i, y_i) = \exp{( \bm{\theta}^t \mathbf{x}_i \, y_i )}
\end{equation}

With these potential functions, the joint probability in \eqref{eq:bag-l} can be rewritten as
\begin{equation}\label{eq:bag-l2}
 P(Y, \mathbf{y} | \mathbf{X}; \bm{\theta}) \propto C^{(Y)}(\sum_i{y_i}) \prod_i{\exp{(\bm{\theta}^t \mathbf{x}_i \, y_i)}}.
\end{equation}
And finally, the bag label likelihood, is obtained by
\begin{equation}\label{eq:bag-label-l}
\begin{split}
 & P(Y | \mathbf{X}; \bm{\theta}) = \sum_{\mathbf{y}}{P(Y, \mathbf{y} | \mathbf{X}; \bm{\theta})} = \frac{Z^{(Y)}}{\sum_{Y'}{Z^{(Y')}}},
\end{split}
\end{equation}
\begin{equation}\label{eq:partition-fcn}
\begin{split}
 &\text{where} \, Z^{(Y)} = \sum_{\mathbf{y}}{\left( C^{(Y)}(\sum_i{y_i}) \prod_i{\exp{(\bm{\theta}^t \mathbf{x}_i \, y_i)}} \right)}
\end{split}
\end{equation}
is the partition function of a standard cardinality potential model, which can be computed efficiently.

In summary, we have a unified probabilistic model that states the probability that a bag (e.g.\ video) receives a label based on classifying individual instances (e.g.\ frames), and a cardinality potential that prefers certain counts of positively labeled instances.

\subsubsection{Parameter Learning}

Since only the bag labels, and not the instance labels, are provided in training, this Cardinality Model is a hidden conditional random field (HCRF).
A commonly used algorithm for parameter learning is maximum a posteriori estimation of the parameters given the parameter prior distributions by maximizing the log likelihood function: 
\begin{equation}\label{eq:MAP-l}
 \mathcal{L}(\bm{\theta}) = \sum_i{\log{P(Y_i | \mathbf{X}_i; \bm{\theta})}} - \lambda \, r(\bm{\theta}).
\end{equation}
This is maximum likelihood optimization of a HCRF with parameter regularization ($r(\bm{\theta}) = \lVert \bm{\theta} \rVert_n$ for $L_n$-norm regularization). Gradient ascent is used to find the optimal parameters, where the gradients are obtained efficiently in terms of marginal probabilities~\cite{quattoni07}.

\subsection{Cardinality Kernel}
\label{subsec:mi-cpk-svm}
This section presents the proposed probabilistic kernel for multi-instance classification.   Kernels operate over a pair of inputs, in this case two bags.  This kernel is defined using the Cardinality Models defined above.  Each bag has its own set of instances, and a probabilistic model is defined over each bag.
A kernel over bags is formed by marginalizing over latent instance labels.

Given two bags $\mathbf{X}_p$ and $\mathbf{X}_q$, a joint kernel is defined between the combined instance features and instance labels for these bags $\mathbf{z}_p = (\mathbf{X}_p, \mathbf{y}_p)$ and $\mathbf{z}_q = (\mathbf{X}_q, \mathbf{y}_q)$:
\begin{equation}\label{eq:joint-kernel}
k_z(\mathbf{z}_p, \mathbf{z}_q) = \sum_{i=1}^{m_p}{\sum_{j=1}^{m_q}{k_x(\mathbf{x}_{pi}, \mathbf{x}_{qj}) k_y(y_{pi}, y_{qj}) } },
\end{equation}
where $k_x(\cdot, \cdot)$ is a standard kernel between single instances, and $k_y(\cdot, \cdot)$ is a kernel defined on discrete instance labels\footnote{If $k_y(\cdot, \cdot)$ is set to 1, the resulting kernel will be equivalent to MI-Kernel~\cite{gartner02}.
Also, note that since the joint kernel is obtained by summing and multiplying the base kernels, it is proved to be a kernel, has all kernel properties, and can be safely plugged into kernel methods.}. By marginalizing the joint kernel w.r.t.\ the hidden instance labels and with independence assumed between the bags, a kernel is defined on the bags as:
\begin{equation}\label{eq:marg-joint-kernel}
\tilde{k}(\mathbf{X}_p, \mathbf{X}_q) = \sum_{\mathbf{y}_p, \mathbf{y}_q}{ P(\mathbf{y}_p | \mathbf{X}_p) P(\mathbf{y}_q | \mathbf{X}_q) k_z(\mathbf{z}_p, \mathbf{z}_q)}.
\end{equation}
Combining the fully observed label instance kernel \eqref{eq:joint-kernel} with the probabilistic version \eqref{eq:marg-joint-kernel}, it can be shown that the marginalized joint kernel is reduced to
\begin{equation}\label{eq:marg-joint-kernel2}
\begin{split}
& \sum_{i=1}^{m_p}\sum_{j=1}^{m_q} \sum_{\mathbf{y}_p, \mathbf{y}_q} \Big( k_x(\mathbf{x}_{pi}, \mathbf{x}_{qj}) k_y(y_{pi}, y_{qj}) \\
& \qquad \qquad \qquad P(y_{pi} | \mathbf{X}_p) P(y_{qj} | \mathbf{X}_q) \Big) .
\end{split}
\end{equation}
In our proposed framework, $P(y_{pi} | \mathbf{X}_p)$ and $P(y_{qj} | \mathbf{X}_q)$ are obtained by
\begin{equation}\label{eq:marg-prob}
P(y_{i} | \mathbf{X}) = \sum_Y{P(y_{i} | Y, \mathbf{X}) \, P(Y | \mathbf{X})},
\end{equation}
where $P(y_{i} | Y, \mathbf{X})$ are the marginal probabilities of a standard cardinality potential model, which can be computed efficiently in $O(m \log^2m )$ time. Also $P(Y | \mathbf{X})$ is the bag label likelihood introduced in \eqref{eq:bag-label-l}.

In general, any kernel for discrete spaces can be used as $k_y$. The most commonly used discrete kernel is $k_y(y_{pi}, y_{qj}) = \mathds{1}(y_{pi} = y_{qj})$. However, since throughout this paper we are dealing with binary instance labels and we are interested in performing recognition with the most salient and positively relevant instances of a bag, $k_y$ is assumed to be
\begin{equation}\label{eq:label-kernel}
k_y(y_{pi}, y_{qj}) = \mathds{1}(y_{pi} = 1) \cdot \mathds{1}(y_{qj} = 1).
\end{equation}
Using this, the kernel in \eqref{eq:marg-joint-kernel2} is simplified as:
\begin{equation}\label{eq:micp-kernel}
\begin{split}
& \tilde{k}(\mathbf{X}_p, \mathbf{X}_q) = \\
& \sum_{i=1}^{m_p}{\sum_{j=1}^{m_q}{  k_x(\mathbf{x}_{pi}, \mathbf{x}_{qj}) P(y_{pi} = 1 | \mathbf{X}_p) P(y_{qj} = 1 | \mathbf{X}_q) } }.
\end{split}
\end{equation}

It is interesting to note that this kernel in \eqref{eq:micp-kernel} can be rewritten as
\begin{equation}\label{eq:micp-kernel-map}
\begin{split}
& \tilde{k}(\mathbf{X}_p, \mathbf{X}_q) = \\
& \Big( \sum_{i=1}^{m_p}{ P(y_{pi} = 1 | \mathbf{X}_p) \Psi(\mathbf{x}_{pi})} \Big) \Big( \sum_{j=1}^{m_q}{ P(y_{qj} = 1 | \mathbf{X}_q) \Psi(\mathbf{x}_{qj})} \Big),
\end{split}
\end{equation}
where $\Psi(\mathbf{x})$ is the mapping function that maps the instances to the underlying feature space of the instance kernel $k_x$. This proves that the unnormalized cardinality kernel in the original feature space corresponds to weighted sum of the instances in the induced feature space of $k_x$, where the weights are the marginal probabilities inferred from the Cardinality Model in the original space. It can be also shown that in the more general case of $k_y(y_{pi}, y_{qj}) = \mathds{1}(y_{pi} = y_{qj})$, the resulting cardinality kernel would correspond to weighted sum of all the instances which take the same instance label in the mapped feature space and concatenating them altogether.

Finally, to avoid bias towards the bags with large numbers of instances, the kernel is normalized as~\cite{gartner02}:
\begin{equation}\label{eq:micp-kernel-norm}
k(\mathbf{X}_p, \mathbf{X}_q) = \frac{\tilde{k}(\mathbf{X}_p, \mathbf{X}_q)}{\sqrt{\tilde{k}(\mathbf{X}_p, \mathbf{X}_p)} \sqrt{\tilde{k}(\mathbf{X}_q, \mathbf{X}_q)}}.
\end{equation}
We call the resulting kernel the \emph{``Cardinality Kernel"}. By using this kernel in the standard kernel SVM, we propose a method for multi-instance classification with cardinality relations.

\subsection{Algorithm Summary}
\label{subsec:summary}
The proposed algorithm is summarized as follows. First the parameters $\bm{\theta}$ of the Cardinality Model are learned (Sec.~\ref{subsec:cpm}). These parameters control the classification of individual instances and the cardinality relations for bag classification.  Next, the marginal probabilities of instance labels under this model are inferred and used in the kernel function in \eqref{eq:micp-kernel}. Finally, the kernel is normalized and plugged into an SVM classifier\footnote{For the parameter setting guidelines, see the supplementary material.}. 

A comprehensive analysis of the computational complexity of the proposed algorithm can be found in the supplementary material. In short, the kernel in \eqref{eq:micp-kernel} can be evaluated in $O(m_p m_q d + m_p \log^2m_p + m_q \log^2m_q)$ time, where the basic kernel $k_x$ takes $O(d)$ time to compute, and the number of instances in each bag are $m_p$ and $m_q$.

\section{Experiments}
\label{sec:exp}

We provide empirical results on three tasks: group activity recognition, video event detection, and video interesting-ness analysis.

\subsection{Collective Activity Recognition}
\label{subsec:activity}
The Collective Activity Dataset~\cite{choi09} comprises 44 videos (about 2500 video frames) of \emph{crossing}, \emph{waiting}, \emph{queuing}, \emph{walking}, and \emph{talking}. Our goal is to classify the collective activity in each frame. To this end, we model the scene as a bag of people represented by the \emph{action context} feature descriptors\footnote{These features are based on a spatio-temporal context region around a person. So by using our cardinality-based model, the spatio-temporal and cardinality information are combined.} developed in \cite{lan12}. We use our proposed algorithms with the ratio-constrained cardinality model in \eqref{eq:majorityPot} with $\rho = 0.5$, to encode a majority cardinality relation. We follow the same experimental settings as used in \cite{lan12}, i.e., the same 1/3 of the video clips were selected for test and the rest for training. The one-versus-all technique was employed for multi-class classification. We applied $l_2$-norm regularization in likelihood maximization of the Cardinality Model and simply used linear kernels as the instance kernels in our method. The results of our Cardinality Kernel are shown in Table~\ref{table:activity-result} and compared with the following methods\footnote{All these methods follow the standard evaluation protocol introduced in~\cite{choi12}. See the supplementary material to find the comparison with the methods in~\cite{amer12, amer13, amer14hirf}, which use a different evaluation setting.}:
(1) SVM on global bag-of-words, (2) Graph-structured latent SVM method in~\cite{lan12}, (3) MI-Kernel~\cite{gartner02}, (4) Cardinality Model of Section~\ref{subsec:cpm} (our own baseline).

\begin{table}[tbh]
\begin{center}
\small
\caption{Comparison of classification accuracies of different algorithms on collective activity dataset. Both multi-class accuracy (MCA) and mean per-class (MPC) accuracy are shown because of class size imbalance.}
\label{table:activity-result}
\begin{tabular}{l|c|c}
\hline
Method &\hspace{-0.05in} MCA\hspace{-0.05in} &\hspace{-0.05in}MPCA \\
\hline
Global bag-of-words with SVM~\cite{lan12} & 70.9 & 68.6\\
Latent SVM with optimized graph~\cite{lan12} & 79.7 & 78.4\\
Cardinality Model & 79.5 & 78.7\\
MI-Kernel & 80.3 & 78.4\\
Cardinality Kernel (our proposed method) & \textbf{83.4} & \textbf{81.9}\\
\hline
\end{tabular}
\end{center}
\end{table}

Our simple Cardinality Model can achieve results comparable to the structure-optimized models by replacing spatial relations with cardinality relations. Further, the proposed Cardinality Kernel can significantly improve classification performance of the Cardinality Model. Finally, our Cardinality Kernel is considerably better than MI-Kernel, showing the advantage of using importance weights (i.e.\ probability of being positive) of each instance for non-uniform aggregation of instance kernels.

\begin{figure}[tbh]
\centering
\subfigure[]{
\centering
\includegraphics[width=0.45\linewidth]{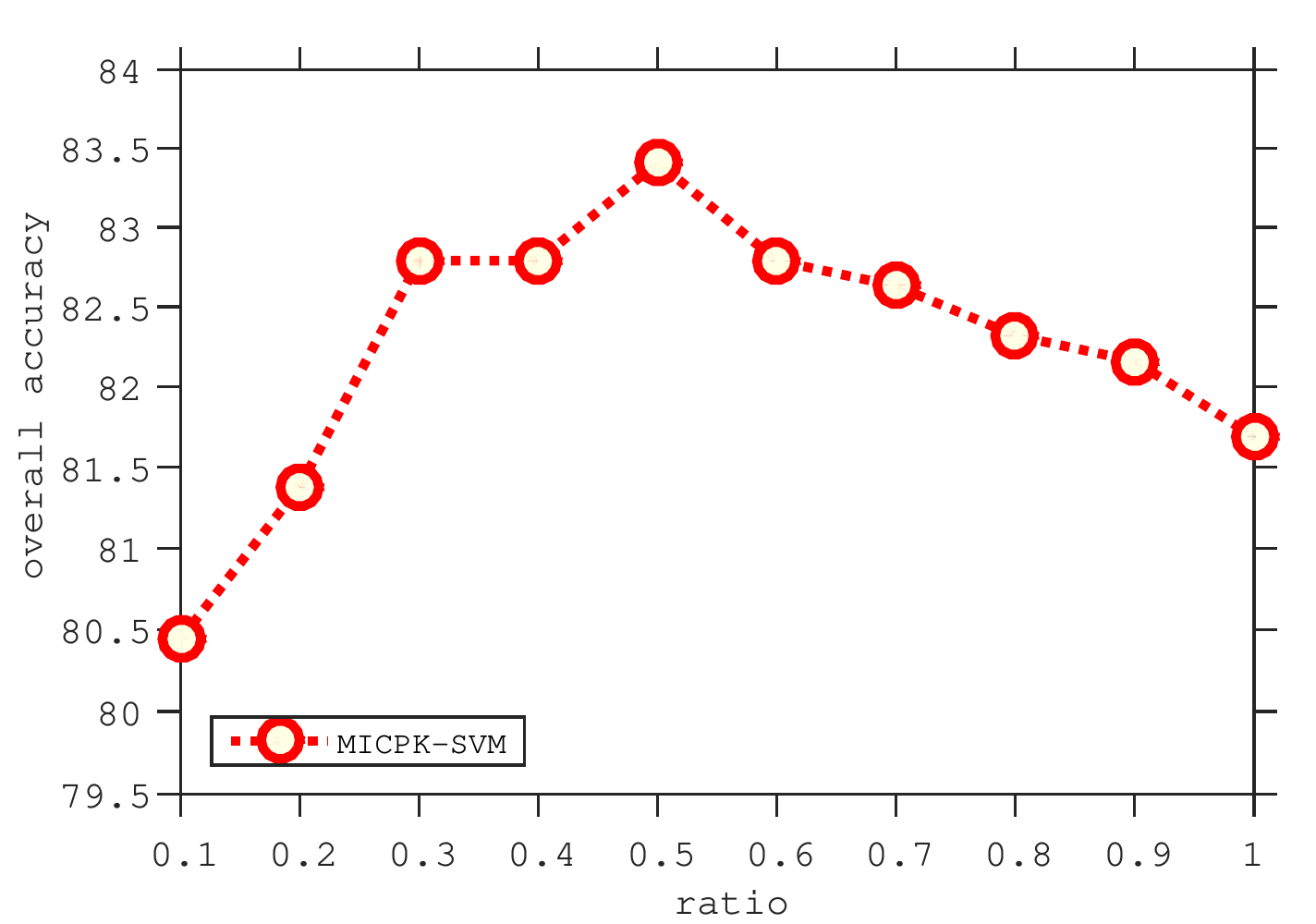}
\label{subfig:actvity-ratios}
}
\subfigure[]{
\includegraphics[width=0.45\linewidth]{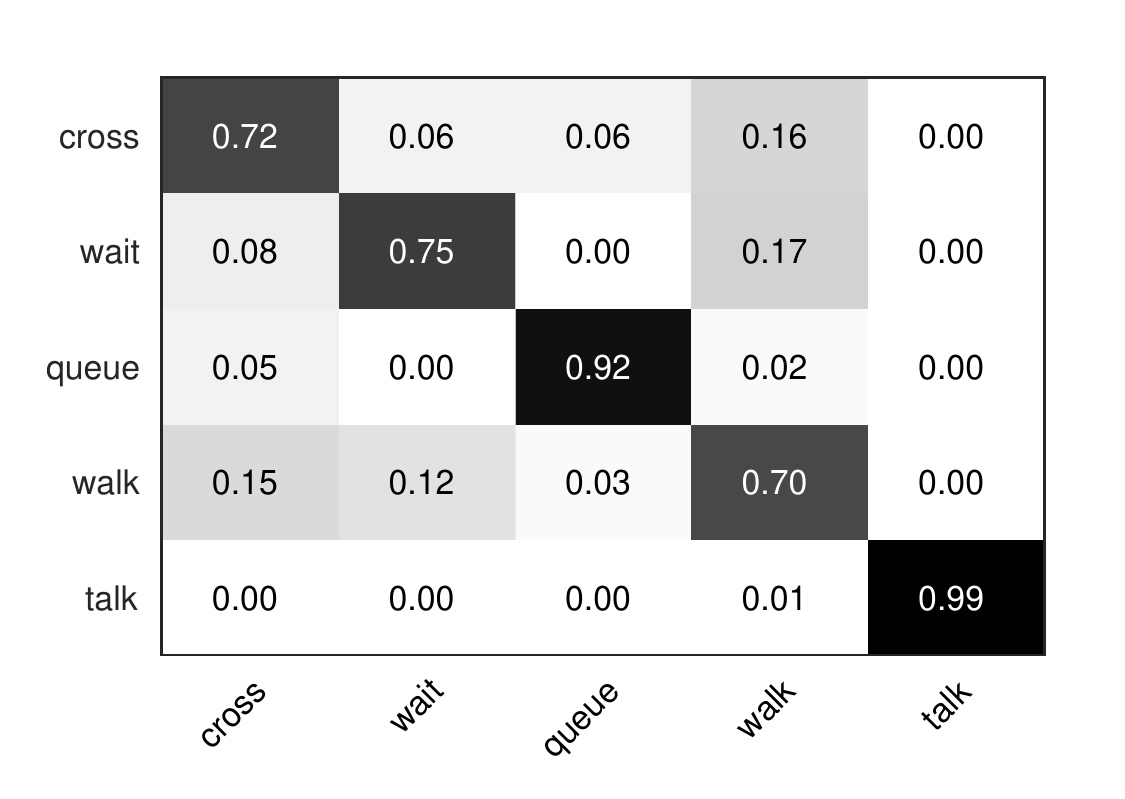}
\label{subfig:activity-confusion-mat}
}
\caption{Performance of the Cardinality Kernel on collective activity dataset. \subref{subfig:actvity-ratios} Classification accuracy with different values of $\rho$ in the ratio-constrained cardinality model. \subref{subfig:activity-confusion-mat} Confusion matrix with $\rho=0.5$ (rows are the true labels, and columns are predicated labels)}
\label{fig:micpk-results}
\end{figure}

\begin{figure}[tbh]
\begin{center}
\includegraphics[width=0.49\linewidth]{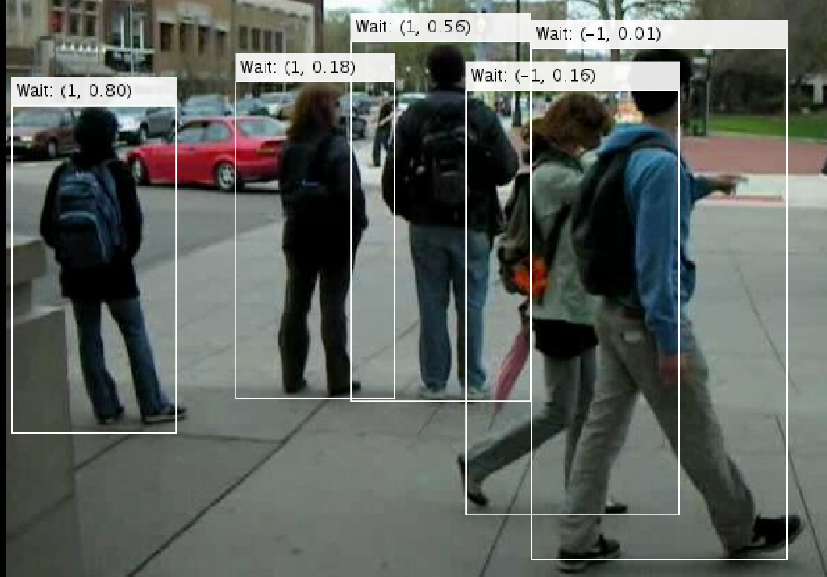}
\includegraphics[width=0.49\linewidth]{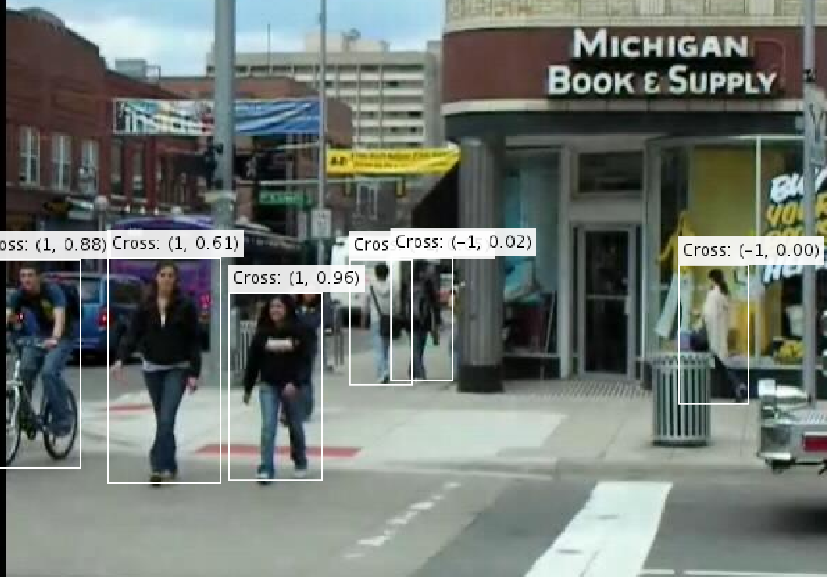}
\end{center}
   \caption{Examples of recognition with the proposed model. The annotation of each person shows the true activity label of the scene with a tuple, indicating the MAP-inferred action label and the corresponding marginal probability w.r.t.~the the scene activity label.  -1 values denote ``not'' of the corresponding category; people performing other actions (left: two people not waiting, right: people not crossing the street) are correctly given -1 labels.}
\label{fig:activity-visual}
\end{figure}

Fig.~\ref{subfig:actvity-ratios} illustrates the effect of $\rho$ in the ratio-constrained cardinality model on classification accuracy of the Cardinality Kernel. It can be seen that as expected, the best result is achieved with $\rho=0.5$. We also provide the confusion matrix for the Cardinality Kernel method in Fig.~\ref{subfig:activity-confusion-mat}. Finally, two examples of recognition with the Cardinality Model for crossing and waiting activities are visualized in Fig.~\ref{fig:activity-visual}.

\subsection{Event Detection}
\label{subsec:event}
We evaluate our proposed method for event detection on the TRECVID MED11 dataset~\cite{over11}. Because of temporal clutter in the videos, not all parts of a video are relevant to the underlying event, and the video segments might have unequal contributions to event detection. Our framework can deal with this temporal ambiguity, i.e., when the evidence of an event is occurring in a video and what the degree of discrimination or importance of each temporal segment is. We represent each video as a bag of ten temporal video segments, where each segment is represented by pooling the features inside it.  As the cardinality potential, we use the Normal model in \eqref{eq:normalPot} with $\mu=1$ and $\sigma=0.1$ to embed a soft and intuitive constraint on the number of positive instances: \emph{the more relevant segments in a video, the higher the probability of occurring the event}.

We follow the evaluation protocol used in~\cite{vahdat13, tang12}. The DEV-T split of MED11 dataset is used for validation and finding the hyper-parameters such as the regularization weights in learning the Cardinality Model and SVM. Then, we evaluate the methods on the DEV-O test collection (32061 videos), containing the events 6 to 15 and a large number of null (or background events). For training, an Event-Kit collection of roughly 150 videos per event is used, and as in~\cite{vahdat13, tang12}, the classifiers are trained for each event versus all the others.

\begin{figure*}[tbh]
\begin{center}
\includegraphics[width=0.95\linewidth, height=1.8in]{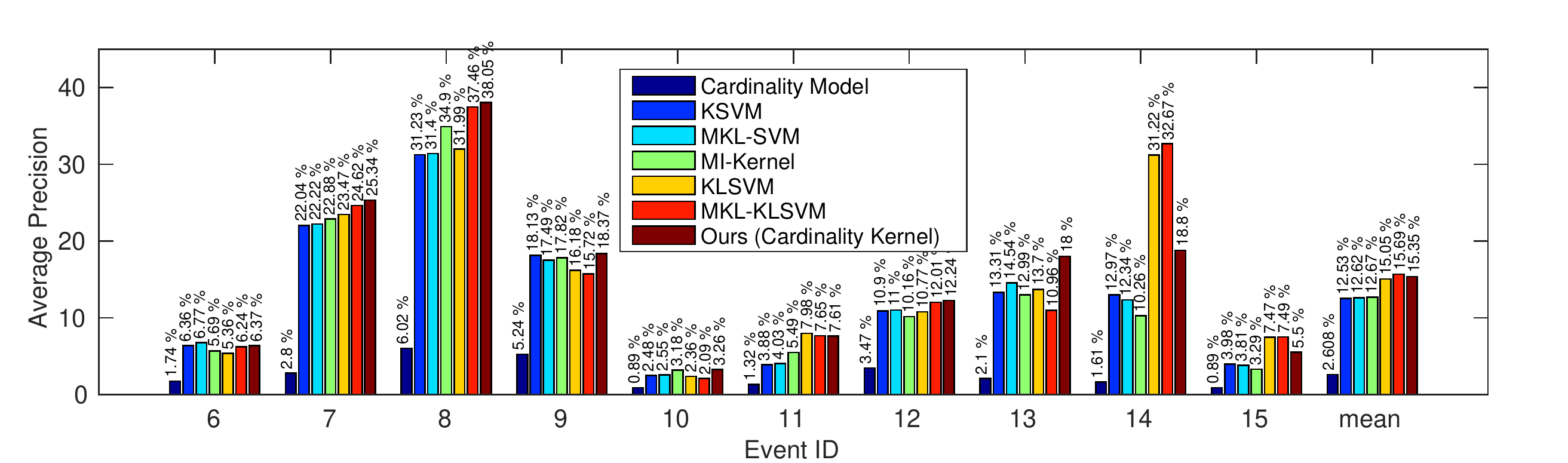}
\end{center}
   \caption{The APs for events 6 to 15 in TRECVID MED 2011. The results for KSVM, MKL-SVM, KLSVM, and MKL-KLSVM are reported from~\cite{vahdat13}. MI-Kernel is based on our own implementation of the algorithm in~\cite{gartner02}.}
\label{fig:events-aps}
\end{figure*}

We compare our methods with the kernelized latent SVM methods in~\cite{vahdat13}, applied to a structured model where the temporal location and scene type of the salient video segments are modeled as latent variables. To have a fair comparison, we use the same set of features: HOG3D, sparse SIFT, dense SIFT, HOG2x2, self-similarity descriptors (ssim), and color histograms, which are simply concatenated to a single feature vector\footnote{In the experiments of this section we compare our method with the most relevant methods, which use the same features. By using, combining, or fusing other sets of features, better results can be achieved (e.g.~\cite{tang13combining, xu14event})}. For training the Cardinality Model regularized maximum likelihood is used with $l_1$-norm regularization, and for the Cardinality Kernel histogram intersection kernel is plugged as the instance kernel. The results in terms of average precision (AP) are shown in Fig.~\ref{fig:events-aps}. It can be observed that based on mean AP, our proposed Cardinality Kernel clearly outperforms the baselines:
\begin{itemize}[leftmargin=*]
\item The Cardinality Model of Sec.~\ref{subsec:cpm}.
\vspace{-0.06in}
\item Kernelized SVM (KSVM) and multiple kernel learning SVM (MKL-SVM), which are kernel methods with global bag-of-words models.
\vspace{-0.06in}
\item MI-Kernel~\cite{gartner02}, which is a multi-instance kernel method with uniform aggregation of the instance kernels.
\end{itemize}
On the other hand, our method is comparable to the kernelized latent SVM (KLSVM) methods in~\cite{vahdat13}. However, our model is considerably less complicated, and unlike these methods, our proposed framework has exact and efficient inference and learning algorithms. For example the training time for our method is about 35 minutes per event, but those methods takes about 30 hours per event\footnote{We performed our experiments on an Intel(R) Core(TM) i7-2600 CPU @ 3.40GHz, and compared to our previous work~\cite{vahdat13}.}. In addition, based on comparison on individual events, our proposed method achieves the best AP in 6 out of 10 events.

Recently, Lai et al.~\cite{lai14} proposed a multi-instance framework for video event detection, by treating a video as a bag of temporal video segments of different granularity. Since this is the closest work to ours, we run another experiment on TRECVID MED11 to evaluate performance of our algorithm compared to~\cite{lai14}. We use exactly the same settings as before, but since Lai et al.~\cite{lai14} used dense SIFT features, we also extract dense SIFT features quantized into a 1500-dimensional bag-of-words vector for each video segment\footnote{We use VLFeat, as in~\cite{lai14}, though with fewer codewords (5000 in \cite{lai14}). See the supplementary material for the results with more codewords.}, where the video segments are given by dividing each video into 10 equal parts. This is slightly different from the multi-granular approach in~\cite{lai14}, where both the single frames and temporal video segments are used as the instances (single--g $\propto$SVM uses only single frames and multi--g $\propto$SVM uses both the single frames and video segments). The results are shown in Table \eqref{table:event-sift-result}. Our method outperforms multi--g $\propto$SVM (which is the best in~\cite{lai14}) by around $20\%$. In addition, our algorithm is more efficient, and training takes only about half an hour per event.

\begin{table}[tbh]
\begin{center}
\small
\caption{Comparing our proposed Cardinality Kernel method with $\propto$SVM algorithms in~\cite{lai14} on TRECVID MED11. The best AP for each event is highlighted in bold}
\label{table:event-sift-result}
\begin{tabular}{l|C{2cm}|C{2cm}|C{1.7cm}}
\hline
Event & single--g $\propto$SVM~\cite{lai14} &multi--g $\propto$SVM~\cite{lai14} & Cardinality Kernel\\
\hline
6 & 1.9 \% & \textbf{3.8} \% & 2.8 \% \\
7 & 2.6 \% & \textbf{5.8} \% & \textbf{5.8} \%\\
8 & 11.5 \% & 11.7 \% & \textbf{17.0} \%\\
9 & 4.9 \% & 5.0 \% & \textbf{8.8} \%\\
10 & 0.8 \% & 0.9 \% & \textbf{1.3} \%\\
11 & 1.8 \% & 2.4 \% & \textbf{3.4} \%\\
12 & 4.8 \% & 5.0 \% & \textbf{10.7} \%\\
13 & 1.7 \% & 2.0 \% & \textbf{4.7} \%\\
14 & 10.5 \% & \textbf{11.0} \% & 4.9 \%\\
15 & 2.5 \% & \textbf{2.5} \% & 1.4 \%\\
\hline
mAP & 4.3 \% & 5.0 \% & \textbf{6.1} \%\\
\hline
\end{tabular}
\end{center}
\end{table}

\vspace{-0.1in}
\subsection{Video Summarization by Detecting Interesting Video Segments}
\label{subsec:vid-sum}
Recently, Gygli et al.~\cite{gygli14} proposed a novel method for creating summaries from user videos by selecting a subset of video segments, which are interesting and informative. For this purpose, they created a benchmark dataset (SumMe\footnote{The dataset and evaluation code for computing the f-measure are available at \url{http://www.vision.ee.ethz.ch/~gyglim/vsum/}}) of 25 raw user videos, summarized and annotated by 15 to 18 human subjects. In their proposed method, each video segment is scored by summing the \emph{interestingness} score of its frames, estimated by a regression model learned from human annotations. At the end, a subset of video segments is selected such that the summary length is $15\%$ of the input video.

\begin{figure}[tbh]
\begin{center}
\includegraphics[width=0.8\linewidth, height=1in]{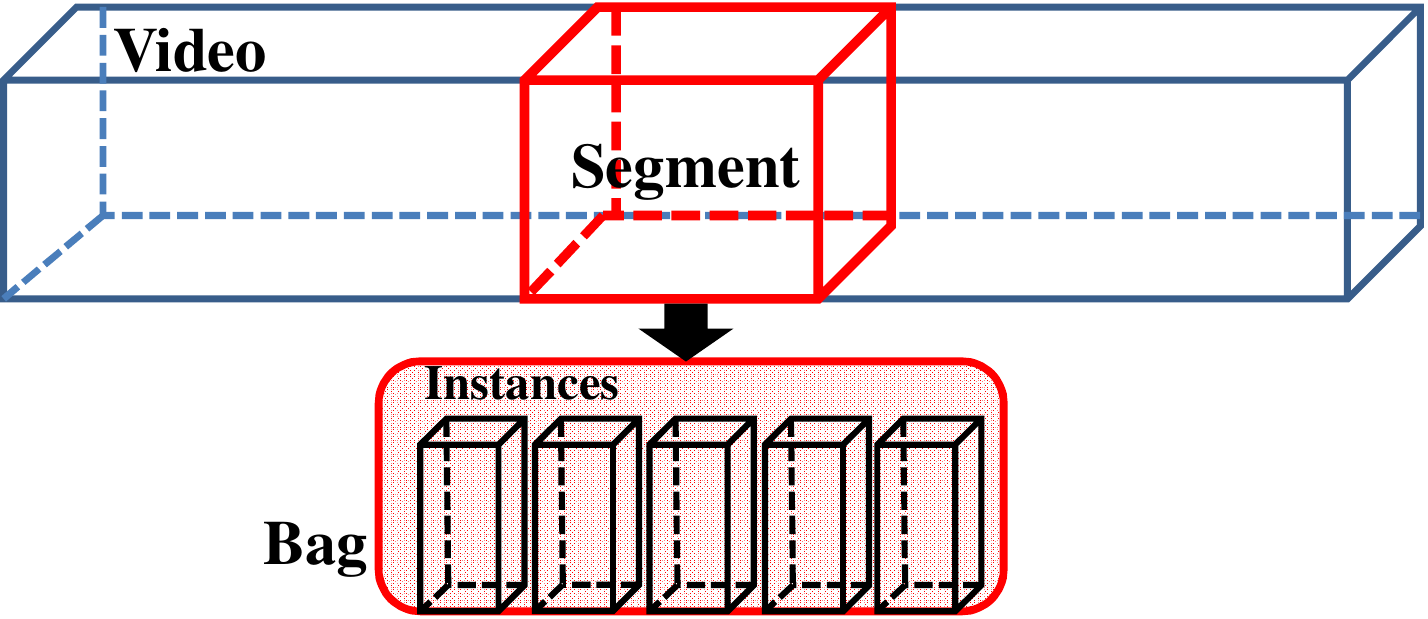}
\end{center}
   \caption{Detecting interesting video segments. A video is modeled as a bag of sub-segments.}
\label{fig:subsegments}
\end{figure}

In this paper, we propose a new approach for creating segment-level summaries. Instead of predicting the per-frame scores and using a heuristic aggregation operation such as ``sum", we use our multi-instance model to directly estimate the interestingness of a video segment. The proposed approach is illustrated in Fig.~\ref{fig:subsegments}. Each segment is modeled as a bag of sub-segments, where a positive bag is a segment which has large overlap with human annotated summaries. To represent each sub-segment, we extract HSV color histogram (with $8 \times 8$ bins) and bag-of-words dense trajectory features~\cite{WangKSL11} (with 4000 words) for each frame and max-pool the features over the sub-segment. Here, we summarize our method and the baselines:
\begin{itemize}[leftmargin=*]
\item Ours: A segment is divided into 5 sub-segments, and the proposed Cardinality Kernel with Normal cardinality potential ($\mu=1, \sigma=0.1$) is used to score the segments.
\item Global Model: A global representation of each segment is constructed by max-pooling the features inside it, and an SVM is trained on the segments.
\item Single-Frame SVM: An SVM is trained on the frames, and the score of each segment is estimated by summing the frame scores.
\item Single-Frame SVR: This is our simulation of the algorithm in~\cite{gygli14} but with our own features, fixed length segments, and using support vector regression.
\end{itemize}
The top scoring $15\%$ of segments are selected in each.

\begin{figure}[tbh]
\begin{center}
\includegraphics[width=0.4\linewidth, angle=90]{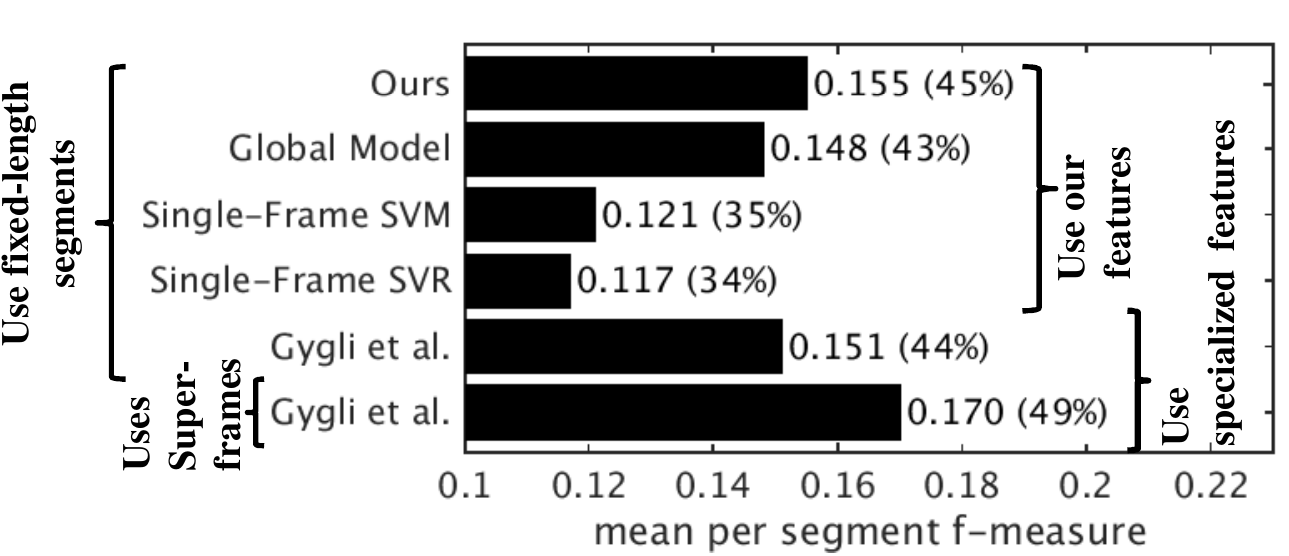}
\end{center}
   \caption{Comparison of different algorithms for segment-level summarization of the SumMe benchmark videos. The percent scores are relative to the average human.}
\label{fig:f-measures}
\end{figure}

For all methods a video is segmented into temporal segments of length $P_l = 1.85$ seconds (the segment length given in~\cite{gygli14}), and histogram intersection kernel is used for training the SVMs.
To evaluate the methods, the procedure in~\cite{gygli14} is used: leave-one-out validation, comparison based on per segment f-measure. The results are shown in Fig.~\ref{fig:f-measures}. It can be observed that our method outperforms the baselines and is competitive with the state-of-the-art results in~\cite{gygli14}. In fact, although we are using general features (color histogram and dense trajectory) we achieve a performance which is comparable to the performance in~\cite{gygli14}, which uses specialized features to represent \emph{attention}, \emph{aesthetics}, \emph{landmarks}, etc. Note that the best f-measure in \cite{gygli14} is obtained by over-segmenting a video into cuttable segments called \emph{superframe}, using guidelines from editing theory.

\section{Conclusion}
\label{sec:conclusion}
We demonstrated the importance of cardinality relations in visual recognition. To this end, a probabilistic structured kernel method was introduced. This method is constructed based on a multi-instance cardinality model, which can explore different levels of ambiguity in instance labels and model different cardinality-based assumptions. We evaluated the performance of the proposed method on three challenging tasks: collective activity recognition, video event detection, and video summarization. The results showed that encoding cardinality relations and using a kernel approach with non-uniform (or probabilistic) aggregation of instances leads to significant improvement of classification performance. Further, the proposed method is powerful, straightforward to implement, with exact inference and learning, and can be simply integrated with off-the-shelf structured learning or kernel learning methods.

{\small
\bibliographystyle{ieee}
\bibliography{cvpr15}
}

\end{document}